\documentclass[11pt,a4paper]{article}
\usepackage[hyperref]{acl2021}
\usepackage{times}
\usepackage{latexsym}
\usepackage{url}
\usepackage{graphicx}
\usepackage{subfigure}
\usepackage{amssymb}
\usepackage{amsmath,bm}
\usepackage{paralist,tabularx,multirow}
\usepackage{caption}
\usepackage{booktabs}

\usepackage{microtype}

\aclfinalcopy 

\title{Multilingual Speech Translation with Unified Transformer:\\ Huawei Noah's Ark Lab at IWSLT 2021}

\author{Xingshan Zeng, Liangyou Li, Qun Liu \\
  Huawei Noah's Ark Lab \\
  \texttt{\{zeng.xingshan,liliangyou,qun.liu\}@huawei.com} \\}

\date{}

\begin{document}
\maketitle
\begin{abstract}
This paper describes the system submitted to the IWSLT 2021 Multilingual Speech Translation (MultiST) task from Huawei Noah's Ark Lab. We use a unified transformer architecture for our MultiST model, so that the data from different modalities (i.e., speech and text) and different tasks (i.e., Speech Recognition, Machine Translation, and Speech Translation) can be exploited to enhance the model's ability. Specifically, speech and text inputs are firstly fed to different feature extractors to extract acoustic and textual features, respectively. Then, these features are processed by a shared encoder--decoder architecture. We apply several training techniques to improve the performance, including multi-task learning, task-level curriculum learning, data augmentation, etc. Our final system achieves significantly better results than bilingual baselines on supervised language pairs and yields reasonable results on zero-shot language pairs.
\end{abstract}

\section{Introduction}
Multilingual Speech Translation (MultiST) aims to develop a single system that can directly translate speech in different languages into text in many other languages. Due to data scarcity of Speech Translation (ST), multilingual and multimodal models are promising as they enable knowledge transferred from other languages and related tasks like Automatic Speech Recognition (ASR) or Neural Machine Translation (NMT). They also allow zero-shot translation in the settings of no direct parallel data.
The IWSLT 2021 MultiST task is held for evaluating the performance under the circumstances. This paper describes our system for the task.

We build a unified model for both speech- and text-related tasks, so that the knowledge from different modalities (speech and text) and different tasks (in this work, the tasks include ST, ASR, and NMT) can be learned together to enhance ST. Specifically, our model consists of three parts -- feature extractor, semantic encoder and decoder. For all the tasks, the semantic encoder and the decoder will be shared to learn unified representations. It follows the Transformer~\cite{DBLP:conf/nips/VaswaniSPUJGKP17} encoder-decoder framework to learn modality-independent features and output text representations. We use the  Conv-Transformer~\cite{DBLP:conf/interspeech/HuangHYC20} as feature extractor for speech input, and the word embedding for text input. The extracted features are then fed to the semantic encoder regardless of the input modality.

However, it is difficult for such a unified model to directly digest knowledge from diverse tasks. Therefore, we apply task-level curriculum learning for our model. We presume the ST task is more difficult than the other two tasks (ASR and NMT), as it not only requires acoustic modeling to extract speech representations, but also needs alignment knowledge between different languages for translation. To this end, our training is divided into three steps -- ASR and NMT pre-training, joint multi-task learning, and ST fine-tuning. What's more, to alleviate the data scarcity problem, we also apply CTC multi-task learning~\cite{DBLP:conf/icml/GravesFGS06}, data augmentation techniques including SpecAugment~\cite{DBLP:journals/corr/abs-1911-08876} and Time Stretch~\cite{DBLP:conf/icassp/NguyenSNW20}, and knowledge distillation~\cite{DBLP:conf/interspeech/LiuXZHWWZ19}, etc.

We conduct experiments in the constrained setting, i.e., only the Multilingual TEDx~\cite{DBLP:journals/corr/abs-2102-01757} dataset is used for training. It contains speech and transcripts from four languages (Spanish, French, Portuguese, and Italian), and some of them are translated into English and/or other languages of the four mentioned above. Several language pairs for ST are provided without parallel training corpus and evaluated as zero-shot translation. The experimental results show that our unified model can achieve competitive results on both supervised settings and zero-shot settings. 

\section{Model Architecture}
\begin{figure}[t]
\centering
\includegraphics[width=72mm]{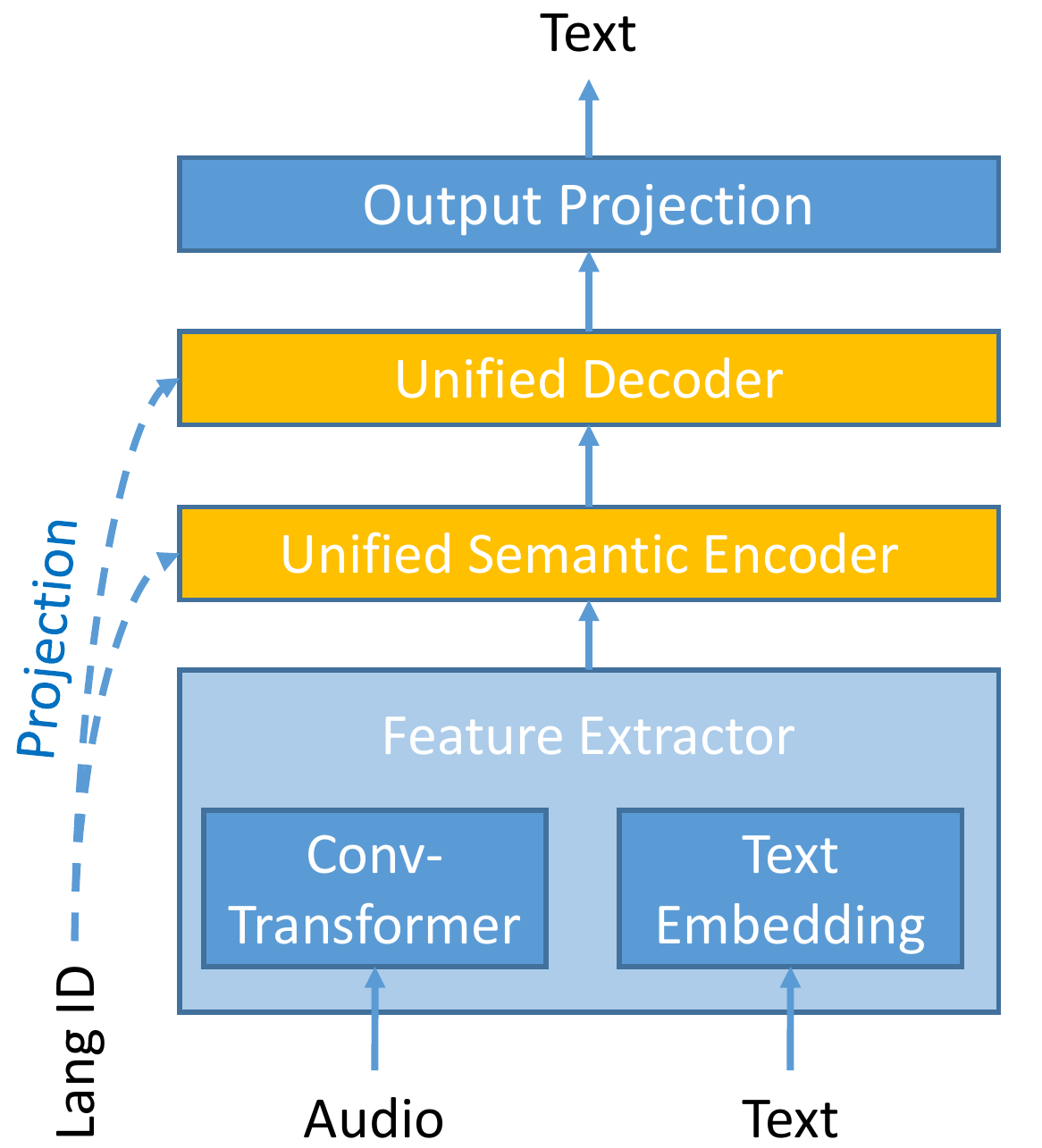}
\vskip -0.5em
\caption{
Overall structure of our unified model. 
}
\vskip -0.5em
\label{fig:model}
\end{figure}
The architecture of our unified model, which is based on Transformer~\cite{DBLP:conf/nips/VaswaniSPUJGKP17}, is shown in Figure~\ref{fig:model}. The NMT part (both input and output is text) follows the basic Transformer setting, i.e. $6$ layers for both the encoder and the decoder, each with $8$ attention heads, $512$ hidden dimensions, and $2048$ hidden units in feed-forward layers. For the speech input, we replace the word embedding layer with the Conv-Transformer~\cite{DBLP:conf/interspeech/HuangHYC20} encoder as acoustic modeling to extract audio features, and the rest are shared. The Conv-Transformer encoder gradually downsamples the speech input with interleaved convolution and Transformer layers. We downsample the speech input $8\times$ times with three Conv-Transformer blocks, each contains three convolution layers (the stride number is $2$ in the second convolution layer, and $1$ in other layers) and two Transformer layers. The Conv-Transformer is set following \citet{DBLP:conf/interspeech/HuangHYC20} and also consistent with the shared parts (in terms of hidden dimensions, etc). Then, the output is fed into the shared semantic encoder and decoder to produce text representations.

For language encoding, we apply language projection~\cite{DBLP:journals/corr/abs-2102-08357} to learn language-specific information. It replaces the language embedding in conventional multilingual models with a projection matrix before the positional embedding layer. With language IDs and input modality, our unified model can recognize the task that needs to be completed. For example, our model will perform ASR with speech input and the same language input and output IDs.

\section{Techniques}
Our model is trained in an end-to-end manner with all available data, including the ASR data (speech and transcript) and the ST data (speech, transcript and translation). From the ST data, we also extract the speech-transcript pairs as ASR data, and the transcript-translation pairs as NMT data. We apply task-level curriculum learning to train our model. At the same time, data augmentation, knowledge distillation, and model ensemble are used to further improve the performance. We describe the techniques in details in the 
rest of this section.

\subsection{Task-Level Curriculum Learning}
\label{subsec:cl}
As a cross-modal and cross-lingual task, ST is more complicated than ASR or NMT. 
Therefore, we presume it is better for our unified model to learn in a smoother way. We divide the training procedure into three steps:
\begin{enumerate}
    \item \textit{ASR and NMT pre-training}: we use all the ASR and NMT data together to pre-train our unified model with a certain number of steps.
    \item \textit{Joint multi-task learning}: all the data including the ST data are used to jointly train the model in a multi-task manner.
    \item \textit{ST fine-tuning}: we fine-tune the model with only ST data to further improve the performance in specific language pairs\footnote{Note that fine-tuning can only be applied in non zero-shot translation language pairs.}.
\end{enumerate}

For all the three steps, we use an additional CTC loss~\cite{DBLP:conf/icml/GravesFGS06} on the output of the last layer of Conv-Transformer encoder to assist with the acoustic modeling. What's more, to make the model focus on the ST task, we assign less loss weights to ASR and NMT tasks (both $0.5$, while $1.0$ for ST) in step 2.

\subsection{Data Augmentation}
We use SpecAugment~\cite{DBLP:conf/interspeech/ParkCZCZCL19,DBLP:journals/corr/abs-1911-08876} and Time Stretch~\cite{DBLP:conf/icassp/NguyenSNW20} to augment the speech data during training.

\paragraph{SpecAugment.} 
SpecAugment is a data augmentation technique originally introduced for ASR, but proven to be effective in ST as well. It operates on the input filterbanks and consists of three kinds of operations, time warping, time masking, and frequency masking. We follow \citet{DBLP:journals/corr/abs-1911-08876} and only apply time masking and frequency masking. It means that a number of consecutive portions of the speech input are masked in the time or the frequency dimensions. We always apply SpecAugment to both the ASR and ST tasks in the three training steps. We set the parameter for time masking $T$ to $40$ and that for frequency masking $F$ to $4$. The number of time and frequency masks applied $m_T$ and $m_F$ are $2$ and $1$, repsectively.

\paragraph{Time Stretch.}
Time stretching is a kind of augmentation method applied in extracted acoustic features like filterbanks to simulate conventional speed perturbation technique~\cite{DBLP:conf/interspeech/KoPPK15}.  Specifically, given a consecutive feature vectors of speech input, it stretches every window of $w$ feature vectors by a factor of $s$ obtained from an uniform distribution of range [$low$, $high$]. In this way, some frames are dropped (if $s>1$) or repeated (if $s<1$) to simulate audio speeding up or down. We only apply Time Stretch in the first two training steps, as we found it does not help much in fine-tuning. We set $w$ to $\infty$, and $low=0.8$, $high=1.25$.

\subsection{Knowledge Distillation}
Teaching the ST model with a pre-trained NMT model using knowledge distillation has been shown effective~\cite{DBLP:conf/interspeech/LiuXZHWWZ19}. Hence we also use word-level knowledge distillation to help with training. Specifically, we minimize the KL divergence between the distribution produced by our model and that produced by the pre-trained NMT model. The tradeoff weight for the knowledge distillation part is set to $0.7$ (i.e. $0.3$ for cross entropy based on ground-truth targets). We use knowledge distillation only in the ST fine-tuning step.

\subsection{Model Ensemble}
Ensemble decoding is to average the word distribution output from diverse models at each decoding step. It is an very effective approach to improve the quality of NMT models. We select the top 2 or 3 models in terms of BLEU scores on development set for each language pair to perform ensemble decoding. The candidate models are trained with different hyper-parameters.

\section{Experiments and Results}

\subsection{Experimental Setup}
We only participate in the constrained setting task. Therefore, only the data from the Multilingual TEDx~\cite{DBLP:journals/corr/abs-2102-01757} is available. It contains speech and transcripts from four languages (Spanish, French, Portuguese, and Italian), and some of them are translated into other languages of the five (English and the four mentioned above). The data statistics are shown in Table~\ref{tab:data}.
\begin{table}[t]
\scriptsize
\begin{center}
\setlength{\tabcolsep}{1.2mm}
\begin{tabular}{c|c c c c c}
\toprule[1pt]
\multirow{2}{*}{\bf{Source}}&\multicolumn{5}{c}{\bf{Target Text}}\\
\cline{2-6}
&\bf{En}&\bf{Es}&\bf{Fr}&\bf{Pt}&\bf{It}\\
\midrule[0.5pt]
\bf{Es}&39k (69h)&107k (189h)&7k (11h)&24k (42h)&6k (11h)\\
\bf{Fr}&33k (50h)&24k (38h)&119k (189h)&16k (25h)&--\\
\bf{Pt}&34k (59h)&$\star$&--&93k (164h)&--\\
\bf{It}&$\star$&$\star$&--&--&53k (107h)\\
\bottomrule[1pt]
\end{tabular} 
\end{center}
\vskip -1.0em
\caption{
The number of sentences and the duration of
audios for the Multilingual TEDx  dataset. Same source and target languages mean the ASR data. Noted with $\star$ are the language pairs for zero-shot translation.
}
\vskip -0.5em
\label{tab:data}
\end{table}

We use 80-dimensional log-mel filterbanks as acoustic features, which are calculated with 25ms window size and 10ms step size and normalized by utterance-level Cepstral Mean and Variance Normalization (CMVN). For transcriptions and translations, SentencePiece\footnote{\url{https://github.com/google/sentencepiece}}~\cite{kudo-richardson-2018-sentencepiece} is used to generate a joint subword vocabulary with the size of 10k. We share the weights for input and output embeddings, as well as the output projection in CTC module.

Our model is trained with 8 NVIDIA Tesla V100 GPUs, each with a batch size of 32. We use Adam optimizer~\cite{DBLP:journals/corr/KingmaB14} during model training with learning rates selected in $\{2e^{-3}, 1e^{-3}, 8e^{-4}, 5e^{-4}, 3e^{-4}\}$ and warm-up steps selected in $\{2000, 6000, 10000\}$, followed by the inverse square root scheduler. Dropout rate is selected in $\{0.1, 0.2, 0.3\}$. We save checkpoints every epoch and average the last 10 checkpoints for evaluation with a beam size of $5$. Our code is
based on fairseq S2T\footnote{\url{https://github.com/pytorch/fairseq/tree/master/examples/speech_to_text}}~\cite{wang-etal-2020-fairseq}.

\begin{table*}[t]
\small
\begin{center}
\setlength{\tabcolsep}{2.05mm}
\begin{tabular}{l|c|c|c|c|c|c|c|c|c|c|c}
\toprule[1pt]
\bf{Model}&\bf{Es-En}&\bf{Es-Fr}&\bf{Es-Pt}&\bf{Es-It}&\bf{Fr-En}&\bf{Fr-Es}&\bf{Fr-Pt}&\bf{Pt-En}&\bf{Pt-Es$^\star$}&\bf{It-En$^\star$}&\bf{It-Es$^\star$}\\
\midrule[0.5pt]
Bilingual&16.60&0.70&16.16&0.50&17.49&13.74&1.26&16.83&--&--&--\\
$\;\;$+ASR data&19.17&9.55&29.59&14.19&24.56&25.13&23.38&21.95&--&--&--\\
\midrule[0.5pt]
Joint learn&23.97&21.76&33.52&22.04&27.65&30.08&30.62&26.36&24.50&14.99&12.34\\
Curriculum learn&25.13&22.72&35.54&24.51&29.75&31.88&31.91&28.07&26.14&15.82&14.98\\
$\;\;$+FT&25.01&22.72&35.04&24.12&29.91&31.87&31.81&27.83&--&--&--\\
$\;\;$+FT with KD&25.25&23.06&35.83&24.68&30.66&32.69&32.96&28.61&--&--&--\\
\midrule[0.5pt]
Ensemble&\bf{26.47}&\bf{23.94}&\bf{36.59}&\bf{25.25}&\bf{31.60}&\bf{33.86}&\bf{34.07}&\bf{29.02}&\bf{27.12}&\bf{16.14}&\bf{16.82}\\
\bottomrule[1pt]
\end{tabular} 
\end{center}
\vskip -1.0em
\caption{
BLEU scores of our unified model for Multilingual TEDx test sets. Those marked with $^\star$ are the results for zero-shot translation. For each setting, we display the results with highest scores among different hyper-parameters. The ensemble results come from ensembling top 2 or 3 models based on the development sets.
}
\vskip -0.5em
\label{tab:main_res}
\end{table*}
\subsection{Results}
This section shows the results of our unified model in Multilingual TEDx dataset. We display the results of our model for MultiST, as well as ASR and NMT, to show the efficacy of our unified model.

\paragraph{MultiST.}
Table~\ref{tab:main_res} shows the results of our model on MultiST. The first two rows display the results with only bilingual data. As can be seen, it is difficult for an end-to-end model to produce reasonable results with extremely low resources (less than 30 hours, including language pairs Es-Fr, Es-It and Fr-Pt as in Table~\ref{tab:data}). With sufficient additional ASR data, all language pairs are improved in a large scale, especially for those low-resource language-pairs (e.g. from $1.26$ to $23.38$ on Fr-Pt).

The rest rows are the results in multilingual settings, where we use all the available data. ``Joint learn'' means that we directly train the multilingual model from scratch. ``Curriculum learn'' displays the results after the first two training steps in Section~\ref{subsec:cl}, while ``+FT'' means adding the third fine-tuning step. ``KD'' refers to knowledge distillation. We can find that ASR and NMT pre-training helps the model learn better representations to perform translation. Then, fine-tuning with knowledge distillation further improve the results. This indicates the efficacy of our task-level curriculum learning for MultiST. However, we find that fine-tuning only with ground-truth targets would not improve the performance. This might be attributed to the limited ST training data, as all of them are less than 100 hours, which introduces difficulty to learn efficiently. By incorporating knowledge distillation, it enables our model to learn extra meaningful knowledge from NMT, so that it can further improve the results.

It can also be found that our unified model can perform reasonable zero-shot speech translation, as all the zero-shot language pairs achieve higher than $10$ BLEU scores. Specifically, results for Pt-Es even achieve similar scores compared with other supervised language pairs. This is mostly because Portuguese and Spanish are similar languages so that it is easier for the model to transfer knowledge from other data.

\begin{table}[t]
\small
\begin{center}
\setlength{\tabcolsep}{2.45mm}
\begin{tabular}{l|c|c|c|c}
\toprule[1pt]
\bf{Model}&\bf{Es}&\bf{Fr}&\bf{Pt}&\bf{It}\\
\midrule[0.5pt]
Monolingual&19.93&22.49&24.86&22.94\\
Multilingual-ASR&13.75&16.79&17.67&16.22\\
Joint learn&15.69&17.46&19.85&19.12\\
Curriculum learn&14.99&16.97&18.06&18.42\\
$\;\;$+FT&\bf{12.53}&\bf{14.56}&\bf{15.75}&\bf{15.38}\\
\bottomrule[1pt]
\end{tabular} 
\end{center}
\vskip -1.0em
\caption{
WER of our unified model for ASR test sets. 
}
\vskip -0.5em
\label{tab:asr_res}
\end{table}
\begin{table}[t]
\small
\begin{center}
\setlength{\tabcolsep}{2.1mm}
\begin{tabular}{l|c|c|c|c}
\toprule[1pt]
\bf{Model}&\bf{Es-En}&\bf{Es-Fr}&\bf{Es-Pt}&\bf{Es-It}\\
\midrule[0.5pt]
Multilingual-NMT&30.41&22.35&41.99&25.62\\
Joint learn&31.11&\bf{28.25}&\bf{44.12}&\bf{27.88}\\
Curriculum learn&30.82&27.87&43.36&27.46\\
$\;\;$+FT&\bf{31.43}&27.81&43.53&27.46\\
\bottomrule[1pt]
\toprule[1pt]
\bf{Model}&\bf{Fr-En}&\bf{Fr-Es}&\bf{Fr-Pt}&\bf{Pt-En}\\
\midrule[0.5pt]
Multilingual-NMT&35.44&36.89&37.46&33.83\\
Joint learn&\bf{37.17}&\bf{39.78}&\bf{40.66}&\bf{35.54}\\
Curriculum learn&36.15&38.83&39.38&34.40\\
$\;\;$+FT&36.42&38.99&39.43&34.78\\
\bottomrule[1pt]
\end{tabular} 
\end{center}
\vskip -1.0em
\caption{
BLEU of our unified model for NMT test sets. 
}
\vskip -0.5em
\label{tab:nmt_res}
\end{table}
\paragraph{ASR and NMT.}
We also test our unified model on the ASR and NMT tasks. Table~\ref{tab:asr_res} and Table~\ref{tab:nmt_res} display the results for ASR and NMT, respectively. ``Multilingual-ASR (NMT)'' is the model trained only with multilingual ASR (NMT) data. From the results, we can find that ASR also benefits from the task-level curriculum learning procedure. However, it only improves slightly compared to the model only with ASR data, probably because the speech in ST data is sampled from the ASR data~\cite{DBLP:journals/corr/abs-2102-01757}. It surprises us that NMT can also benefit from extra data from different modality (i.e. speech), although curriculum learning does not improve the performance (probably because we assign less loss weight to NMT task in step 2 as introduced in Section~\ref{subsec:cl}).
This demonstrates the potential of leveraging data from different modalities to train a powerful unified model. Due to the time and data constraint, we leave the exploration into a more powerful unified model with multiple kinds of data as future work.

\paragraph{Submissions.}
We submit our results on ST evaluation sets with the ensemble model in Table~\ref{tab:main_res}, scoring BLEU scores $35.4$ on Es-En, $27.0$ on Es-Fr, $43.2$ on Es-Pt, $30.8$ on Es-It, $26.7$ on Fr-En, $27.0$ on Fr-Es, $26.9$ on Fr-Pt, $26.7$ on Pt-En, $27.0$ on Pt-Es, $17.6$ on It-En, and $15.4$ on It-Es. We also submit our ASR results on evaluation sets with our fine-tuned model (i.e. ``+FT'' model in Table~\ref{tab:asr_res}), scoring $11.1$ WER on Es ASR, $22.2$ on Fr 
ASR, $16.2$ on It ASR, and $23.8$ on Pt ASR.

\section{Conclusions}
We present our system submitted to IWSLT 2021 for multilingual speech translation task. In our system, we build a unified transformer-based model to learn the knowledge from different kinds of data. We introduce a task-level curriculum learning procedure to enable our unified model to be trained efficiently. Our results show the efficacy of our unified model to perform multilingual speech translation in both supervised settings and zero-shot settings. Moreover, the results demonstrate the potential of incorporating multilingual and even multi-modal data into one powerful unified model.

\bibliographystyle{acl_natbib}
\bibliography{anthology,acl2021}


\end{document}